\documentclass[mathematics,article,accept,pdftex,moreauthors]{Definitions/mdpi} 

\usepackage{amsfonts}
\usepackage{physics} 
\usepackage{gensymb}
\usepackage{bbding}
\usepackage{mathtools}
\usepackage{centernot} 
\usepackage{bigints}
\usepackage{dsfont} 
\usepackage{esint} 
\usepackage{multirow}
\usepackage{dsfont}
\usepackage{bbm}
\usepackage{subcaption}

\usepackage{algorithm}
\usepackage{algpseudocode}

\newcommand{\methodname}{AMED}

\usepackage{textcomp}
\usepackage{gensymb}
\newcommand{\bb}[1]{\bm{\mathrm{#1}}}

\usepackage{bm}

\firstpage{1} 
\pubvolume{1}
\issuenum{1}
\articlenumber{0}
\pubyear{2024}
\copyrightyear{2024}
\externaleditor{Academic Editor: Xiang Li}
\datereceived{7 April 2024 } 
\daterevised{28 May 2024  } 
\dateaccepted{4 June 2024  } 
\datepublished{ } 
\hreflink{https://doi.org/} 

\Title{{AMED:} Automatic Mixed-Precision Quantization for Edge~Devices}

\TitleCitation{AMED: Automatic Mixed-Precision Quantization for Edge Devices}

\Author{Moshe 
 Kimhi *\orcidA{}, Tal Rozen, Avi Mendelson and Chaim Baskin}

\AuthorNames{Moshe Kimhi, Tal Rozen, Avi Mendelson and Chaim Baskin}

\AuthorCitation{Kimhi, M.; Rozen, T.;  Mendelson, A.; Baskin, C.}

\address[1]{
{Computer Science department, Technion 
 IIT,} Haifa 3200003, Israel; mendlson@technion.ac.il (A.M.); chaimbaskin@technion.ac.il (C.B.), }

\corres{\hangafter=1 \hangindent=1.05em \hspace{-0.82em} Correspondence: moshekimhi@campus.technion.ac.il}

\abstract{Quantized neural networks are well known for reducing the latency, power consumption, and model size without significant harm to the performance. This makes them highly appropriate for systems with limited resources and low power capacity.
Mixed-precision quantization offers better utilization of customized hardware that supports arithmetic operations at different bitwidths.
Quantization methods either aim to minimize the compression loss given a desired reduction or optimize a dependent variable for a specified property of the model (such as FLOPs or model size); both make the performance inefficient when deployed on specific hardware, but more importantly, quantization methods assume that the loss manifold holds a global minimum for a quantized model that copes with the global minimum of the full precision counterpart. 
Challenging this assumption, we argue that the optimal minimum changes as the precision changes, and thus, it is better to look at quantization as a random process, placing the foundation for a different approach to quantize neural networks, which, during the training procedure,  quantizes the model to a different precision, looks at the bit allocation as a Markov Decision Process, and then, finds an optimal bitwidth allocation for measuring specified behaviors on a specific device via direct signals from the particular hardware architecture. By doing so, we avoid the basic assumption that the loss behaves the same way for a quantized model. Automatic Mixed-Precision Quantization for Edge Devices (dubbed \methodname) demonstrates its superiority over current state-of-the-art schemes in terms of the trade-off between neural network accuracy and hardware efficiency, backed by a comprehensive evaluation.
}


\begin{document}



\section{Introduction}
\label{sec:intro}

Deep neural networks have established themselves as the primary algorithmic solution for a wide array of real-world applications. However, the~computational and memory requirements of DNNs are considerable, leading to notable latency and power consumption during both the training and inference~processes.

To meet the rapidly increasing demand for algorithmic capabilities in embedded systems, such as autonomous vehicles, drones, and~medical devices, prior research has suggested various techniques for reducing these devices' power and energy consumption.
Some of the techniques already deployed in different systems are matrix compression~\cite{Lebedev2015SpeedingupCN, Ullrich2017SoftWF,9206968,JMLR:v22:20-1374}, pruning~\cite{Han2015LearningBW, Wen2016LearningSS}, hardware-aware neural architecture search~\cite{Liu2019DARTSDA, Wu2019FBNetHE, Cai2019ProxylessNASDN}, and~quantization~\cite{zhou2016dorefa,hubara2017,Choi2018PACTPC}.

Quantization is a promising and straightforward technique to accelerate neural network architectures because it reduces the model's memory footprint and the computation complexity by a large~factor.

The standard metric to evaluate a quantized model's computational complexity is the number of multiply--accumulate (MAC) operations or bit operations (BOPs) it demands. Neither consider other factors such as communication complexity~\cite{9177369}, memory utilization, and~inter-layer relations.
Using ultra-low bit quantization does not, therefore, consistently improve the chip's overall performance due to communication and memory boundaries~\cite{su13020717}.

To this end, recent developments in industry~\cite{apple,nvidia} and academia~\cite{samajdar2020systematic,Sharma2018BitFB} have introduced  support for various precision matrix multiplications, leading to a new line of work focusing on mixed-precision (MP) quantization~\cite{Wang2018HAQHA,Dong2019HAWQHA,Sun2022FILMQNNEF}, that is assigning a different bandwidth to each matrix (i.e., weights and activations).

\vspace{-2pt}\begin{figure}[H]
	\includegraphics[width=1.\textwidth]{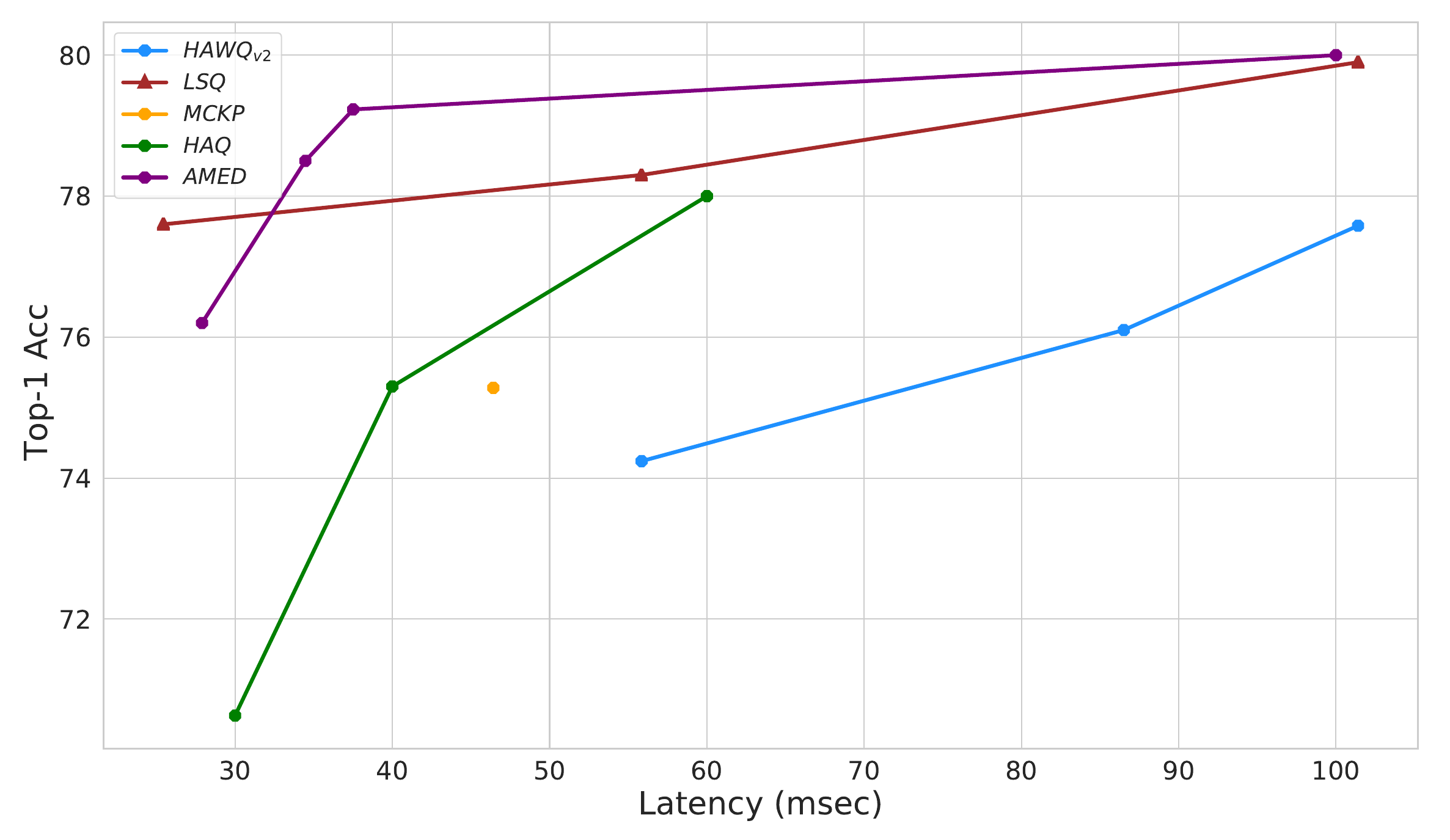} 
	\caption{{ResNet-50} 
 quantized models on a latency--accuracy plane. Circles are mixed-precision quantization, and triangles are uniform quantization.
	Our models achieve a better Pareto curve of dominant solutions in the two-dimensional plane for ultra-low precision.}
	\label{fig:resnet50}
\end{figure}

MP quantization methods focus on either finding the optimal solution by the estimation of the loss manifold of a full-precision model and assess what layers would cause the least steepest change to the local minimum the model reached at one quantization step~\cite{Dong2019HAWQHA,Wang2018HAQHA}. In~this work, we argue that, when quantizing a model {\textit{during}} 
 the training process, the minimum on this loss manifold could be very different. This realization led us to the conclusion that the quantization should be looked at as a process, and~the optimal step depends on the intermediate state of the model, not on a final form. This way, an~intermediate quantized model would be easier to train and the quantization loss would be lower, so the trajectory to the final quantized model is~smoother.

Hardware-aware methodologies, sometimes coupled with Neural Architectural Search (NAS), have shown promising progress in the field of hardware-aware models. Current work~\cite{Wu2019FBNetHE,Cai2019ProxylessNASDN} directly measures signals from the hardware simulator. Nevertheless, to~incorporate hardware constraints into the loss, they neglect the dependencies between layers, that is the~latency per layer per precision is a fixed number. This assumption is inaccurate in cases where the memory utilization is high because the communication boundary agrees with 
the computational one. NAS can provide a solution that is better tailored to the task. Even so, the~search space is commonly huge, and~the cost of training~\cite{Sun2019AnEL} and the carbon footprint~\cite{Strubell2019EnergyAP} are~high.

Our solution, Automatic Mixed-Precision Quantization for Edge Devices (dubbed \methodname), is an algorithmic framework that chooses mixed-precision bit allocation per layer by looking at reduced precision as a Markov Decision Process (MDP), using signals directly from the hardware. 
The procedure is simple to use, not bounded to a specific HW design, and easily fine-tuned by the user to find a good balance between performance and resource~consumption, as shown in \cref{fig:resnet50}.

This paper makes the following contributions:

\begin{itemize}
    \item A novel framework for mixed-precision quantization for DNNs that look at reduced precision as a Markov process.
    \item A quality score that represents the accuracy--latency trade-off with respect to the hardware constraint. This allows us to create custom-fit solutions for a range of device-specific hardware constraints via direct hardware signals in the training procedure.
    \item Extensive experiments conducted on different hardware setups with different models on standard image classification benchmarks (CIFAR100, ImageNet). These outperform previous methods in terms of the accuracy--latency trade-off.
    \item A proposed modular framework, i.e.,~the sampling method, hardware properties, accelerator simulator, and~neural network architecture are all independent modules, making it applicable to any given case.  
\end{itemize}

Our source code, experimental settings, and quantized models are available at

 {\url{https://github.com/RamorayDrake/AMED/}.}


\section{Materials}
\label{sec:related}
\unskip
\subsection*{Multi-Objective Optimization}
\label{subsec:mul obj opt}
In a simple learning paradigm, the~optimization process attempts to minimize an objective function; the example in this work is a classification model, minimizing the cross-entropy loss, denoted by $\mathcal{L}_{CE}$.

Multi-objective problems can have both positively correlated objectives, as well as negatively correlated objectives (also known as conflicting objectives or adversary objectives).
The problem we aim to solve in this paper is the accuracy--latency trade-off of the quantized model, in~which one objective is minimizing $\mathcal{L}_{CE}$, while the other is minimizing the latency, denoted by $\mathcal{L}_{lat}$. This idea can be extended to other hardware objectives (such as model size or power consumption); however, these typically correlate with $\mathcal{L}_{lat}$ and are beyond the scope of this~work.  

Both objectives are conflicting because quantizing our model to a lower representation aims to reduce the latency while increasing the cost of 
quantization error.
Due to these conflicting objectives, multi-objective optimization problems are known to lack optimal solutions with respect to all of the objectives~\cite{10.1162/evco.1994.2.3.221}. Typically, there are many optimal (or suboptimal) solutions because one is not comparable with the other.
{Solution} 
 {$\textbf{A}$} dominates solution $\textbf{B}$ ($\textbf{A} \prec \textbf{B}$) if all objective values of $\textbf{A}$ are better than or equal to the respective objective value of $\textbf{B}$.
Dominant solutions define a Pareto-optimal curve in the objective's plane.
To solve multi-objective optimization, it is very common to use single-objective approaches or a Pareto~approach.

A {single-objective approach} 
 aggregates the objectives into one term, where the basic aggregation method is the weighted sum vector~\cite{kdeb01}, denoted by $\alpha$, which encodes the user's priorities over the different objectives. This method is very common when training a DNN, but~has the following~drawbacks:
\begin{itemize}
\item The weighted sum vector $\alpha$ must be determined beforehand and~requires a grid search or meta-learning, both of which can be costly and have difficulty converging.
\item Not all objectives can be optimized via the same optimization scheme. For~example, a~gradient-based optimizer cannot be used when only some of the objectives are~differentiable.
\end{itemize}

A Pareto approach is based on sampling methods and finding the Pareto curve by adopting only dominant solutions.
The main drawback of this approach in the perspective of a quantization scheme is the computational cost. Each sample needs to first quantize a neural network and train it, before~it can evaluate the~objectives.

\section{Literature~Review}
\unskip
\subsection*{Quantized Neural Networks}
Quantization is one of the most efficient and commonly used techniques for the acceleration and compression of models. Generally, quantization procedures can be categorized into Post-Training Quantization (PTQ) and Quantization-Aware Training (QAT): 
\begin{itemize}
    \item \textbf{PTQ} uses a small calibration set to obtain the optimal quantization parameters without the need or capability to use the entire dataset.
    \item \textbf{QAT} conducts the quantization during the training process and, thus, uses the entire training corpus.
   \end{itemize}
   
QAT is more appealing when quantizing a model into ultra-low precision because the network is trained to withhold quantization noise~\cite{Li2017TrainingQN}.
Because we are focusing on ultra-low precision, we use the QAT technique for this~work. 

Homogeneous quantization quantizes both weights and activations using the same bitwidth throughout the network, scaling from 32-bit to as low as binary~\cite{Hubara2016BinarizedNN,Rozen2022BimodalDB}. \mbox{{Ref.} \cite{Zhang2018LQNetsLQ}} proposes a learnable quantizer with a basis vector that is adaptable to the weights and activations. {Refs.}~\cite{Choi2018PACTPC,Baskin2021NICENI} presents a clipping method that automatically optimizes the quantization scales during model training. In~\cite{Esser2020LSQ}, a~non-uniform step size was learned as a model parameter so that it became more sensitive to the quantization transition points. {Ref.}~\cite{Han2021ImprovingLN} takes this method a step further and proposes a weight regularization algorithm that encourages a sharp distribution for each quantization bin and encourages it to be as close to the target quantized value as possible. {Ref.}~\cite{Gong2019DifferentiableSQ} employed a series of hyperbolic tangent functions to approach the staircase function for low-bit quantization~gradually.

Recent research, however, has shown that different layers in a DNN model contribute to the performance differently and contain different redundancies. Therefore, the~notion of a mixed-precision quantization scheme that assigns a different precision bitwidth to each layer of the DNN can offer better accuracy while compressing the model even further. Nevertheless, determining each layer's precision is quite challenging because of the large search space. In~\cite{Zur2019TowardsLO}, NAS was used for allocating the bitwidth by employing BOPs in the cost function. HAQ~\cite{Wang2018HAQHA} leverages reinforcement learning to determine the quantization policy layerwise. Additionally, it takes the hardware accelerator's feedback into account in the architecture design, attaining an optimized solution for deterministic constraints. It fails, however, to~find a solution that stochastically reduces latency and power consumption. FBNET~\cite{Wu2019FBNetHE} combines all possible design choices into a stochastic super-net and approximates the optimal scheme via sampling. These search methods can require a large quantity of computational resources, which scale up quickly with the number of~layers.

Therefore, other works attempt to allocate the bitwidth using different methods. {Ref.}~\cite{Zhao2021DistributionawareAM} used the first-order Taylor expansion to evaluate the loss sensitivity due to the quantization of each channel, and~then adjusted the bitwidth channelwise. {Ref.}~\cite{Yang2021BSQEB} considered each layer's bitwidth as an independent trainable parameter. {Ref.}~\cite{Yang2021FracBitsMP} allocated differentiable bitwidths to layers, which consequently offers smooth transitions between neighboring quantization bit levels, all while meeting a target computational constraint. HAWQ~\cite{Dong2019HAWQHA,Dong2020HAWQV2HA} applies mixed-precision quantization using the Hessian information. Power iteration is adopted to compute the top Hessian eigenvalue, which is used to determine which layers are more prone to quantization. It relies on a proxy signal that assumes that all devices benefit from a higher compression rate and a lower number of operations, and~does not consider the specific hardware design. {Ref.}~\cite{Chen2021TowardsMQ} also used the Hessian trace and formulated the mixed-precision quantization as a discrete constrained optimization problem solved by a greedy search algorithm. {Ref.}~\cite{Zhang2021DifferentiableDQ} differentially learned the quantization parameters for each layer, including the bitwidth, quantization level, and~dynamic range. 
FILM-QNN~\cite{Sun2022FILMQNNEF} investigates mixed-precision quantization in the layer scope. This means that each layer's parameters have a different precision. While this has the benefit of being even more efficient, the~metadata of which parameters are sent to which multiply and accumulate (MAC) unit are not feasible in most current hardware architectures.

The quantization can also be performed in two different ways. Works such \linebreak as~\mbox{\cite{Esser2020LSQ,Gong2019DifferentiableSQ,Yang2021BSQEB,Wang2018HAQHA,Dong2020HAWQV2HA}} perform uniform quantization in which the width of the quantization bins is a single parameter. A~more complex non-uniform quantization allows different bin widths, which reduces the quantization error~\cite{Zhang2018LQNetsLQ,Zhao2021DistributionawareAM}. 

To perform a standard MAC operation in a fixed-point representation on a given hardware, all of the data (weights and activations) must be equally scaled. This means that, when implementing non-uniform quantization, a~lookup table to a higher representation is required. While the communication costs are reduced, the~computational and area costs of non-uniform quantization are very high and inefficient.
\section{Method}
\label{sec:method}
This section introduces our proposed technique for mixed-precision quantization. It leverages the strengths of both approaches described in Section~\ref{subsec:mul obj opt}. We achieve this by first combining the desired properties of the model into a single, unified objective function. Then, we employ the Pareto optimization approach to efficiently sample solutions that achieve a desirable trade-off between these properties.
We define the problem (Section~\ref{method:problem def}) and describe our perspective of the quantization process of neural networks (Section \ref{method" MMC}), following which, we describe our MP quantizer (Section \ref{method: quantizer}) and our MP simulator (Section~\ref{method: sim}). Finally, we introduce the technique of \methodname{} (Section \ref{method: AMED}).

\subsection{Problem~Definition}
\label{method:problem def}
\begin{figure}[H]
	\includegraphics[width=.9\textwidth]{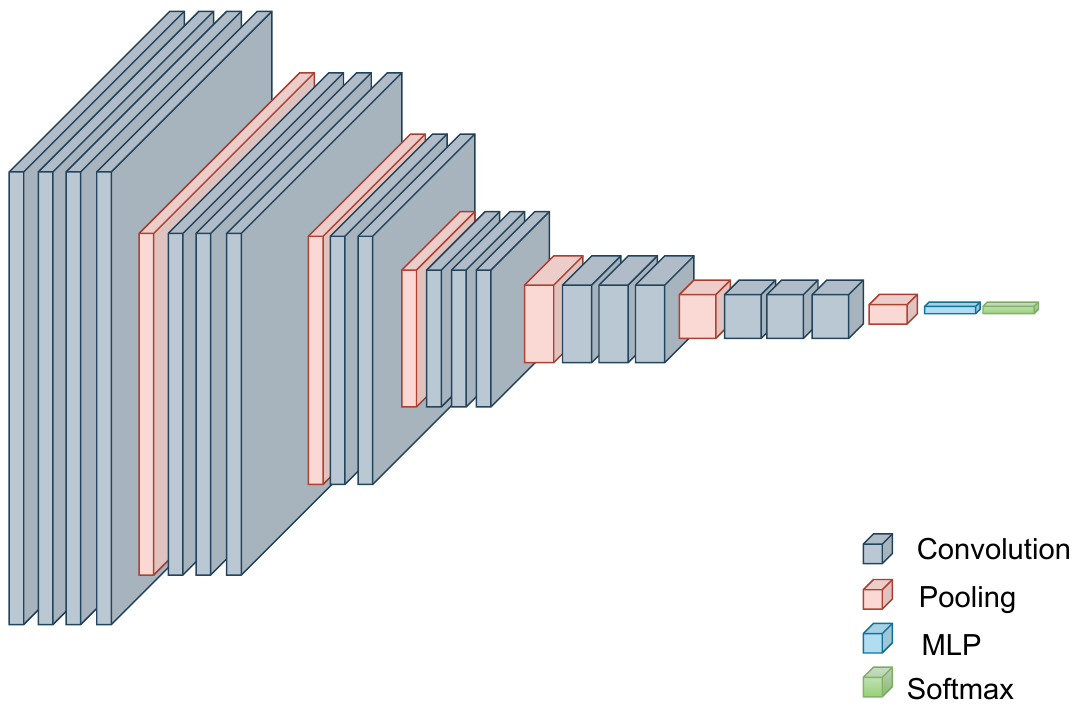} 
	\caption{{A diagram} 
 of DNN architecture activation~maps.}
	\label{fig2}
\end{figure}

The DNN architecture parameters grouped into $\bb{N}$ layers are denoted by ${\theta}$ {(Figure \ref{fig2}).} 
 The~bit-allocation vector is denoted by $\bb{A} \in \mathbb{M}^{\bb{N}}$, where $\mathbb{M}$ is the number of possible bit allocations. Additionally, a~quantized set of model parameters is denoted by ${\theta_A}$.
$\bb{A}$ is a temporal vector, denoted by $\bb{A}_t$ at time $t$, and~the $l$-th layer bitwidth at time $t$ is denoted by $\bb{A}^l_t$. 

\vspace{-2pt}

Due to the dependence on specific accelerator properties, a~differentiable form of the hardware objective (e.g., latency) cannot be directly measured in advance. This renders traditional optimization techniques relying on gradient descent inapplicable. Furthermore, the~non-i.i.d. nature of latency across layers adds another layer of complexity. Quantization choices for one layer (e.g., $\bb{A}^{l-1}$) can influence the cache behavior of subsequent layers (e.g., l), leading to unpredictable changes in the overall latency. This makes individual layer optimization~ineffective.



Motivated by multi-objective optimization approaches that can handle trade-offs between competing goals, we propose encoding user-specified properties of the DNN model (e.g., accuracy, memory footprint, and~latency) into a single, non-differentiable objective function. By~doing so, we aim to avoid introducing relaxations that might not accurately capture the true behavior of the hardware platform, potentially leading to suboptimal~solutions.

{We define:} 
\begin{align} \label{eq:Z}
\mathbf{Z} =  \mathcal{L}_{CE} + \beta\mathcal{L}_{Lat}
\end{align}

Since the objective function $\mathbf{Z}$ is non-differentiable with respect to the model parameters, we employ a nested optimization approach. This approach involves optimizing the inner loop (minimizing $\mathcal{L}_{CE}$) before tackling the outer loop (minimizing $\mathbf{Z}$).

To identify a set of solutions that best balance the trade-off between model performance and hardware constraints, we introduce two modifications to our objective function:

\begin{itemize}
\item We introduce a penalty term that becomes increasingly negative as the quantized model's accuracy falls below a user-specified threshold. This ensures that solutions prioritize models meeting the desired accuracy level.
\item We incorporate a hard constraint on the model size. This constraint acts as a filter, preventing the algorithm from exploring solutions that exceed a user-defined maximum size for the quantized model.
\end{itemize}



With respect to the above and by using a maximization problem, our objective is defined as~follows:
\begin{align} \label{eq:objective2}
\bb{max} \: \: \{(\mathbf{Z}_{ref}-\mathbf{Z}) \odot M\}
\end{align}
where:
$$M=\begin{cases}
\bb{1} , & \text{size} < \text{Memory size} \\
0, & else
\end{cases}$$
and $\mathbf{Z}_{ref}$ would be the baseline performance.
We used a uniform 8-bit quantized network performance as $\mathbf{Z}_{ref}$.

\subsection{Multivariate Markov~Chain}
\label{method" MMC}

Quantization error contributes to the overall approximation error. These errors can be conceptualized as non-orthogonal signals within the error space. Crucially, the~error vectors may not always point in the same~direction.

Simply quantizing the model's weight matrices and adding the resulting error to predictions is insufficient. Furthermore, quantized models without a subsequent fine-tuning step often~underperform.

Nahshan et al.~\cite{Nahshan2021LossAP} demonstrates that the quantization error signals of models with similar bitwidths exhibit smaller angles in the error space, indicating higher similarity. This suggests that, even for models destined for very low precision, an~intermediate quantization step (e.g., 8 bits) can be~beneficial.

To the best of our knowledge, the~optimal approach for progressively quantizing deep learning models, especially those with mixed precision and complex architectures, remains an open question. The~high dimensionality of such a process makes traditional ablation studies~challenging.

While it is well known that optimization of DNNs is not an MDP nor canonical for each precision~\cite{duchi2011adaptive,kingma2014adam}, Nahshan observations motivate our exploration of temporal precision in the quantization process. We propose modeling the network's state as a function of its previous mixed-precision configuration, akin to a multivariate Markov~chain.

Formally, we define the bit allocation as a multivariate Markov chain, as~defined in~\cite{Ching2007OnMM}.

Let ${A^{(i)}}_t$ be the bit allocation of the $i$-th layer at time t, defined as:
\begin{align} \label{eq:MC_one_enterance}
{A^{(i)}}_{t+1} = \lambda_{ii} \mathbf{P}^{(ii)} {A^{(i)}}_{t} + \sum_{j=1,j \ne i}^N \lambda_{ij} {A^{(j)}}_{t} \text{ for} \: \: i=1,2,\dots,N.
\end{align}
where:
\begin{align} \label{eq:MC_lambdas}
\lambda_{ij} \ge 0,\: \sum_{j=1}^N \lambda_{ij} = 1 \: \: \text{for} \: \: i=1,2,\dots,N.
\end{align}
and $\mathbf{P}^{(ii)}$ is a one-step transition probability matrix for the $i$-th layer precision as a Markov~chain.

In the context of this work, every layer of the neural network $i$ has a precision at time $t$, and~the transition for it to a new precision at $t+1$ is given by \cref{eq:MC_one_enterance}.

We denote the multivariate transition matrix of all layers as $\mathcal{Q}$:
\begin{align} \label{eq:transition_matrix_Q}
\mathcal{Q} = \begin{pmatrix}
\lambda_{11} \mathbf{P}^{(11)} & \lambda_{12}I & \dots & \lambda_{1N}I\\
\lambda_{21}I & \lambda_{22} \mathbf{P}^{(22)} & \dots & \lambda_{2N}I \\
\vdots & \vdots & \ddots & \vdots \\
\lambda_{N1}I & \lambda_{N2}I & \dots & \lambda_{NN} \mathbf{P}^{(NN)}
\end{pmatrix}  
\end{align}
and thus, the~bit-allocation multivariate Markov chain update rule is:
\begin{align} \label{eq:A_update}
\bb{A}_{t+1} = \mathcal{Q} \bb{A}_{t}
\end{align}

Modeling $\mathcal{Q}$ explicitly is challenging, since we would like to model it so close minima of the loss will have the highest probability.
As an alternative to explicit modeling, we suggest using sampling techniques over $\hat{\mathcal{Q}}$ where $\mathcal{Q} \propto \hat{\mathcal{Q}}$.
In this study, we use random walk Metropolis–Hastings~\cite{Metropolis1953}, a~Markov chain Monte Carlo (MCMC) method.
For $\hat{\mathcal{Q}}$, we construct a distribution from Objective \ref{eq:objective2}.

We use an exponential moving average (EMA), denoted by $\gamma$, to~avoid cases of diverged samples of quantized models:
\begin{align} \label{eq:Q_hat}
\hat{\mathcal{Q}}_{t+1} = \gamma \hat{\mathcal{Q}}_t +  \mathbf{q}
\end{align}
where  $\gamma = 0.01$ empirically reduces the number of required~samples.

The update of new samples $\mathbf{q}$ is by applying a logarithmic scale over Equation~(\ref{eq:objective2}):
\begin{align} \label{eq:q}
\mathbf{q} = \log( \frac{\mathbf{Z}_{ref}}{\mathbf{Z}}) \odot \bb{M} = \log( \frac{\mathcal{L}_{CE}^{ref} +\beta\mathcal{L}_{Lat}^{ref}}{ \mathcal{L}_{CE} + \beta\mathcal{L}_{Lat}}) \odot \bb{M}
\end{align}
where $\mathbf{q}$ is the update step of the transition matrix $\hat{\mathcal{Q}}$, which is simply updating the probability of transitioning to a different precision based on the scaled loss. $\mathbf{Z}$  is defined in Equation~(\ref{eq:Z}) by the weighted sum of losses. Note that the only difference here from \cref{eq:objective2} is the monotonic logarithm function, as~well as taking the mask out of the log. since the mask operated elementwise (Hadamard product) with the~objective. 

We employ the random walk Metropolis--Hastings algorithm (Algorithm~\ref{alg:rwmh}) to determine whether to accept a new bit-allocation vector $\bb{A}$ or retain the current~one.

\begin{algorithm}[H]
   \caption{{Random} 
 walk Metropolis–Hastings~step.}
   \label{alg:rwmh}
\begin{algorithmic}
   \State {\bfseries Input:} $\hat{\mathcal{Q}}$,  $\bb{A}_i$
   \State  $bb{A}_*$ = $\arg\max \hat{\mathcal{Q}}$ \Comment{Axis 2}
   \State $\alpha = \frac{\hat{\mathcal{Q}}[\bb{A}_*]}{\hat{\mathcal{Q}}[\bb{A}]}$ \Comment{$\alpha$ is the layerwise acceptance ratio};
   \If{0  $ \leq \alpha \leq 1$} \Comment{Element-wise}
   \State $\bb{B} \sim \text{Bern}(\alpha)$
   \State $\bb{A}_{i+1} = \bb{A}_* \bb{B} + \bb{A}_i (\bb{1}-\bb{B})$ 
       \Else
    \State $\bb{A}_{i+1} = \bb{A}_*$
    \EndIf

\end{algorithmic}
\end{algorithm}

The algorithm proceeds in the following steps:

\begin{enumerate}
  \item \textbf{Candidate Generation:} A candidate vector, denoted by $A_*$, is proposed for the next~allocation.
  \item \textbf{Layerwise acceptance ratio:} A layerwise acceptance ratio, $\alpha$, is~calculated.
  \item \textbf{Bernoulli-Based Acceptance:}
     \begin{itemize}
       \item If $\alpha \leq 1$, a~Bernoulli distribution with probability $\alpha$ is used for~acceptance:
           \begin{itemize}
             \item If the Bernoulli trial succeeds, $A_*$ is accepted.
             \item Otherwise, the~current allocation is retained.
           \end{itemize}
       \item If $\alpha \geq 1$, $A_*$ is automatically accepted.
     \end{itemize}
\end{enumerate}

This acceptance scheme adaptively balances exploration and exploitation:

\textbf{Exploration Phase:} Accepts a higher proportion of new candidates, encouraging broad space exploration.
\textbf{Exploitation Phase:} Preferentially accepts candidates that leverage knowledge from previously discovered~samples.


\subsection{Quantizer}
\label{method: quantizer}
Our approach is agnostic to the specific quantization technique employed. As~long as the chosen quantizer is compatible with the user's target accelerator, it can be seamlessly integrated. The~experiments presented in this paper leverage a quantization-aware training (QAT) quantizer along with a learnable quantization scale, as~advocated in~\cite{Esser2020LSQ}.


During the optimization of the quantized network, we employ a technique called fake quantization. This involves maintaining a full-precision (FP32) copy of the weights, while using a simple straight-through estimator for back-propagation~\cite{Bengio2013EstimatingOP}. This estimator approximates the gradients of the quantized weights during~training.

To quantize a matrix $M$ with $b$ bits, we follow a two-step process:
\begin{enumerate}

\item \textit{Scaling and rounding:}
   $$M_{int} = \text{round}\left( \frac{M}{S} \right)$$
   {Here,} 
 $M_{int}$ represents the integer version of $M$, obtained by scaling $M$ by a factor $S$ (often referred to as the scaling factor) and then rounding the~result.

\item \textit{Clamping:}
   $$\bar{M} = \text{clamp}(M_{int}, \min, \max)$$
   The scaled and rounded integer $M_{int}$ is then clamped to the valid range representable by $b$ bits. This ensures the quantized values stay within the intended range.
   \begin{itemize}
   \item[-] {For} 
 symmetric quantization, $\min = -2^{(b-1)}+1$ and $\max = 2^{(b-1)}-1$.
   \item[-] For asymmetric quantization, $\min = 0$ and $\max = 2^{(b-1)}$.
   \end{itemize}
\end{enumerate}

\begin{figure}[H]
	\includegraphics[width=.95\textwidth]{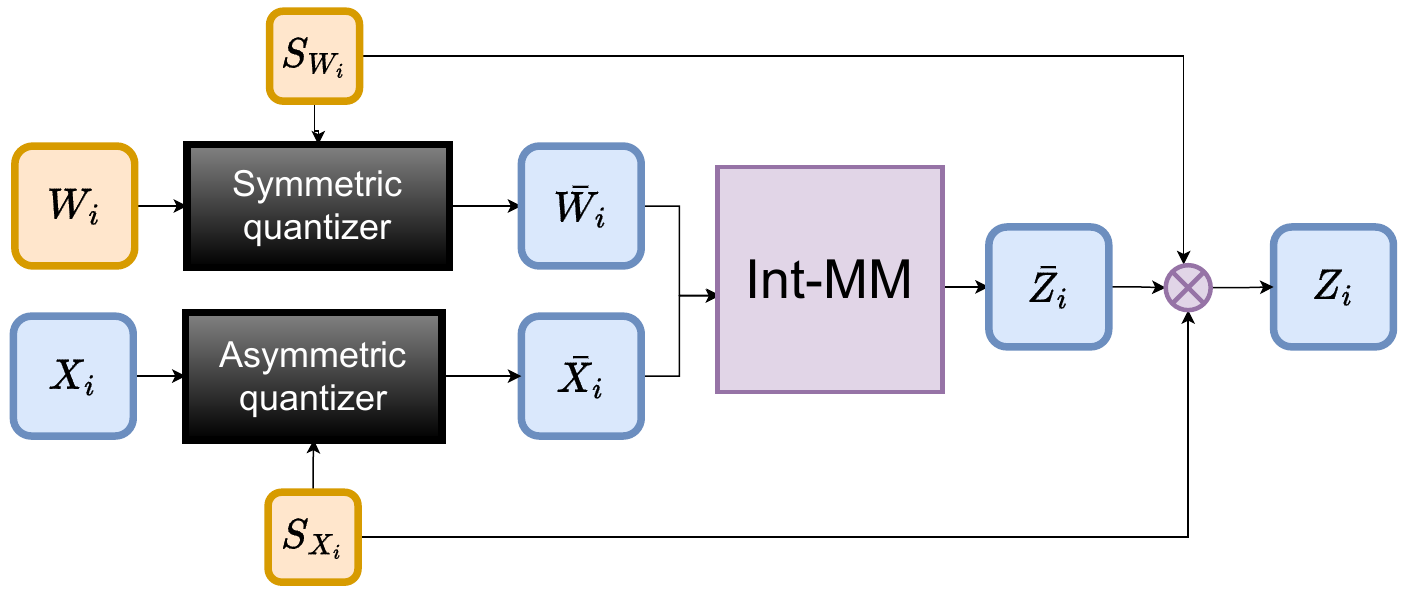} 
	\caption{Quantization of the $i$-th layer of the network. $W_i$ indicates the weights, $X_i$  the input activations, and $\Bar{W_i}, \Bar{X_i}$ the quantized versions, respectively. $S_{W_i},S_{X_i}$ are the learnable parameters of the quantization. (scale). In~Blue are dynamically changed tensors, while in orange the parameters. Weights and activations are quantized with respect to the scaling factor (with rounding and clamping as described in \cref{method: quantizer}. The~quantized versions are multiplied in an integer matrix multiplication accelerator and produce a quantized vector $\Bar{Z_i}$. With respect to the scaling factors, we can dequantize them 
into $Z_i$, which can yield a prediction in FP, or quantize them 
again in a different precision in the next~layer. }
	\label{fig:quan}
\end{figure}

The clamping values depend on the chosen quantization scheme: symmetric or asymmetric. A~figure illustrating low-bit multiplication in a layer is provided in {Figure}
 \ref{fig:quan}.

\subsection{Simulator}
\label{method: sim}

To enable direct sampling of signals from emulated hardware, we developed a hardware accelerator simulator. This simulator allows us to model various aspects of an accelerator, including the inference time for different architectures, and~generate signals usable within our algorithm. These signals can extend beyond latency to encompass memory utilization, energy consumption, or~other relevant~metrics.

\subsubsection{{Underlying}  
 Architecture Modeled}
Our simulator draws heavily on {SCALE-Sim} v2~\cite{samajdar2020systematic}, a~Systolic Accelerator Simulator (SAS) capable of cycle-accurate timing analysis.
A SAS is ideal for DNN computations due to its efficient operand movement and high compute density. This setup minimizes global data movement, largely keeping data transfer local (neighbor to neighbor within the array), which improves both energy efficiency and speed.
Our SAS can additionally provide power/energy consumption, memory bandwidth usage, and~trace results, all tailored to a specific accelerator configuration and neural network architecture. We extended its capabilities by incorporating support for diverse bitwidths and convolutional neural network (CNN) architectures not originally supported, such as Inverted Residuals (described in~\cite{Sandler2018MobileNetV2IR}).

\subsubsection{{Simulator Approximations}}

1. \texttt{{Optimal} 
 Data Flow Assumptions}: The simulator models specific types of dataflows---Output Stationary (OS), Weight Stationary (WS), or Input Stationary (IS)---and assumes an ideal scenario where outputs can be transferred out of the compute array without stalling the compute operations. In~real-world implementations, such smooth operations might not always be feasible, potentially leading to a higher actual~runtime.

2. \texttt{Memory Interaction}: This simplistically models the memory hierarchy, assuming a double-buffered setup to hide memory access latencies. This model may not fully capture the complex interactions and potential bottlenecks of real  memory~systems.

The original \textit{ScaleSim} simulator was validated against real hardware setups using a detailed in-house RTL model~\cite{samajdar2020systematic}.



\subsubsection{{Using the Simulator}}
The simulator operates on two key inputs:

1. {{Network} 
 Architecture Topology File:} This file specifies the structure of the neural network, including the arrangement of layers and their~connections.

2. {{Accelerator Properties Descriptor:}} This descriptor defines the characteristics of the target hardware accelerator, such as its memory configuration and processing~capabilities.

We evaluated the simulator using various network architectures: ResNet-18, ResNet-50~\cite{He2016DeepRL}, and~MobileNetV2~\cite{Sandler2018MobileNetV2IR}. While MobileNetV3 could potentially be explored in future work, it is not included in the current set of experiments. 
The~specific accelerator properties used are detailed in Table~\ref{tab:HW setup}.

\begin{table}[H]\setlength{\tabcolsep}{2.05mm}
    \caption{The accelerator setup for a compact accelerator based on {SCALE-Sim}~\cite{MAX78000}, the~{SCALE-Sim} micro-controller with low memory, and Eyeriss~\cite{Chen2017EyerissAE}.
    The proprieties we used in the simulator for each setup are listed in the table.
    Data flow indicates the stationarity (weights---``ws''; activations---``as''; output---``os''), i.e.,~what data should remain in the SRAM for the next computed layer. ``os''  writes the output directly to the input feature map SRAM.}
    \label{tab:HW setup}
    \begin{tabular}{c c c c}
    \toprule
    \textbf{Name} & \textbf{{SCALE-Sim}} & \textbf{{SCALE-Sim} Low Mem}  & \textbf{Eyeriss v1} \\
    \midrule
    PE array height & 32 & 32 & 12\\
    PE array width & 32 & 32 & 14\\
    Input feature map SRAM (KB) & 64& 4 & 108\\
    Filter SRAM (KB) &64 & 4 & 108\\
    Output feature map SRAM (KB) &64 & 4 & 108\\
    Data flow & os & os & ws\\ 
    Bandwidth (w/c) & 10 & 10 & 10\\ 
    Memory banks & 1 & 1 & 1\\ 
    Speed (GHz) & 0.2 & 0.1 & 0.2\\ 
    \bottomrule
    \end{tabular}

\end{table}

\textbf{SRAM Utilization Estimation:}

The current simulator estimates SRAM utilization based on bandwidth limitations. Incorporating a more accurate calculation of SRAM utilization within the simulator is a potential area for future improvement. This would provide a more precise signal for the~algorithm.

\textbf{Simulator Output:}

The simulator generates a report for each layer, containing various metrics such as the following:
\begin{itemize}
\item[-] {Compute cycles;} 
\item[-]Average bandwidths for DRAM accesses (input feature map, filters, output feature~map);
\item[-] Stall cycles;
\item[-] Memory utilization (potentially improved in future work);
\item[-] Other details specified in Appendix~\ref{app:reports}.
\end{itemize}
\textbf{Extracting Latency Metrics:}

From the reported compute cycles and clock speed ($f$), we calculate the computation latency as $\frac{C}{f}$. Similarly, the~memory latency for each SRAM is estimated using:

$$ \frac{b}{M-BW \times \text{word size}} \times f $$
where
\begin{itemize}
\item[-] {$C$} 
 denotes the compute cycles;
\item[-] $f$ denotes the clock speed;
\item[-] $b$ denotes the number of bits required for the specific SRAM;
\item[-]$M-BW$ denotes the memory bandwidth;
\item[-] Word size is assumed to be 16~bits.
\end{itemize}

The total latency of the quantized model is determined by the maximum latency value obtained from these calculations (computation and memory latencies for each layer).


\subsection{Training and Quantizing with \methodname{}}
\label{method: AMED}


\methodname{} employs a Metropolis--Hastings algorithm (Algorithm~\ref{alg:rwmh}) to sample precision vectors $\bb{A}$. These precision vectors guide the quantization process, resulting in a mixed-precision model ${\theta_A}$. Details on the quantization procedure, including activation quantization matching weight precision, can be found in Section~\ref{method: quantizer}.

Following quantization, we perform a two-epoch optimization step using stochastic gradient descent (SGD) to fine-tune both the quantized model parameters ${\theta_A}$ and the scaling factors $S_{W_i}$ and $S_{X_i}$ associated with the weights and activations, respectively.

The quantized model's performance is evaluated on the validation set using the cross-entropy loss $\mathcal{L}_{CE}^l$ and on a simulated inference scenario detailed in Section~\ref{method: sim}, using the latency loss $\mathcal{L}_{lat}^l$. Both loss values contribute to updating the estimated expected utility $\hat{\mathcal{Q}}$ (Equation~(\ref{eq:Q_hat})) and guide the sampling of a new precision vector $\bb{A}$.

\begin{figure}[H]
	\includegraphics[width=1.\textwidth]{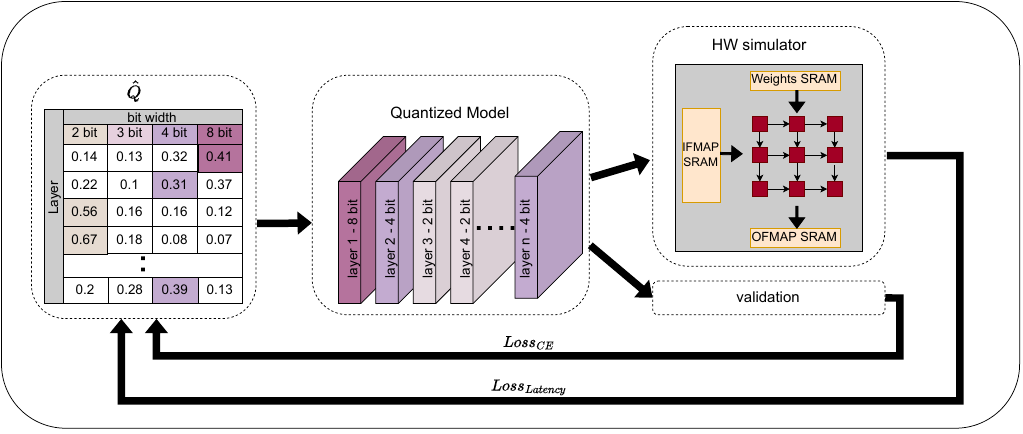} 
	\caption{An illustration of Algorithm \ref{alg:algo}. $\hat{\mathcal{Q}}$ Table represents bit-allocation vector $\bb{A}$  
}
	\label{fig:diagram}
\end{figure}

Figure~\ref{fig:diagram} illustrates the quantization process, while Algorithm~\ref{alg:algo} details the complete~workflow.

\begin{algorithm}
   \caption{Training procedure of \methodname{}.}
   \label{alg:algo}
\begin{algorithmic}
   \State {\bfseries Input:} dataset: $\bb{D}(x,y)$, model: ${\theta}$, params: $\beta, \gamma$, simulator $\bb{S}$
   \State $\bb{A}^i_0 = 8 ; \forall i \in L$ 
   \State ${\theta}_0 ={\theta}_{\bb{A}_0}$ 
   \State $\hat{\mathcal{Q}} \sim U(2,8) $ \Comment{between all bit representations}
   \State Fit ${\theta}_0$
   \State Compute reference $\hat{\mathcal{L}_{CE}}({\theta}_0,\bb{D})$,$\hat{\mathcal{L}_{Lat}}(\bb{S},{\theta}_0)$
   \For{$i$ in epoch}
   \State Evaluate $\mathcal{L}_{CE}({\theta}_i,\bb{D})$,$\mathcal{L}_{Lat}(\bb{S},{\theta}_i)$
   \State Update $\hat{\mathcal{Q}}$ \Comment{from (\ref{eq:Q_hat})}
   \State Update $\bb{A}_i$ \Comment{by Algorithm \ref{alg:rwmh}}
   \State Quantize the model $\theta_{\bb{A}_i}$
   \State Fit ${\theta}_{\bb{A}_i}$
   \EndFor
\end{algorithmic}
\end{algorithm}

The algorithm described in Algorithm \ref{alg:algo} is agnostic to the quantizer and to the hardware specification and simulation. This means that one can use any quantization technique that relies on the statistics of the weights and activations in a single layer and any hardware or hardware simulator and~use \methodname{} to choose the best mixed-precision bit allocation for the~hardware.




\section{Results}
\label{sec:experiments}
In this section, we present our quantization experimental results on the ImageNet dataset~\cite{deng2009imagenet} for different~architectures. 

We applied our \methodname{} algorithm to determine the bitwidth of each layer. We used an SGD optimizer with a momentum of 0.9 and a weight decay of $10^{-3}$ for ResNet and~a weight decay of $10^{-5}$ for MobileNet. We ran each network for 80--90 epochs with a starting learning rate of $10^{-2}$ for ResNet and $10^{-3}$ for MobileNet. The~learning rate dropped by a factor of 10 every 30 epochs. The~batch size was 256, and~we used common data augmentations of random horizontal flip and random crop.  The values of $\beta$ as listed in Table~\ref{tab:imagenet} are $\beta_1 = 1$ and $\beta_2 = 10$.
We also followed the common practice of not quantizing the classifier (FC layer), which is less than 3\% of the latency of the smallest network we trained.
All experiments used a pretrained model quantized uniformly to 8 bits by the regime of~\cite{Esser2020LSQ}.
We applied a uniform quantization technique and learnable scale for both weights and activations, as~described in Figure~\ref{fig:quan}. The~scale factor $S^W$ for the weights was initialized with the statistics from the INT8 quantized model, and~$S^X$ for activations was initialized with the statistics from one batch (of size 256) in the following form:
\begin{align} \label{eq:initilize S}
S^x_0 = \frac{max(\|X \|)^2} { 2^{(b-1)}-1}
\end{align}
where X is the first batch of images and~$b$ is the number of bits that represent the quantized layer.
All experiments using the {SCALE-Sim} accelerator hardware setup for the ImageNet dataset are listed in Table~\ref{tab:imagenet}. Other  hardware performance parameters are given in Appendix~\ref{tab:imagenet hw3}.

For comparison, we tested the quantization performance of our method, as well as other methods found in the literature: HAWQ~\cite{Dong2020HAWQV2HA}, MCKP~\cite{Chen2021TowardsMQ}, LSQ~\cite{Esser2020LSQ}, DDQ~\cite{Zhang2021DifferentiableDQ}, and LIMPQ~\cite{tang2022mixedprecision} for ResNet-18, HAQ~\cite{Wang2018HAQHA}, HAWQ~\cite{Dong2020HAWQV2HA}, MCKP~\cite{Chen2021TowardsMQ}, LSQ~\cite{Esser2020LSQ}, and LIMPQ~\cite{tang2022mixedprecision} for ResNet-50, and MCKP~\cite{Chen2021TowardsMQ}, DDQ~\cite{Zhang2021DifferentiableDQ}, HAQ~\cite{Wang2018HAQHA}, PROFIT~\cite{Park2020PROFITAN}, and LSQ + BR~\cite{Han2021ImprovingLN} for~MobileNet-V2.

\begin{table}[H]
    \caption{{Performance} 
 comparison with state-of-the-art methods
on ImageNet, noticeable good results in bold. $N_{MP}$ indicates mixed precision using $N$ as the minimum allowed bitwidth.
$\psi$ is our re-implementation, with~pretrained FP32 weights from~\cite{rw2019timm}. We only present our implementation when we achieved better results than the original paper. If~the original paper we are comparing to did not publish the model's bit allocation, we could not run the simulator and find the latency or calculate the model size. }
    \label{tab:imagenet}
    \begin{tabular}{c l@{\hspace{0.75cm}} c@{\hspace{0.55cm}} c c c c}
    \toprule
    \textbf{Network} & \textbf{Method} & \textbf{Bitwidth} & \textbf{Acc (\%)} & \textbf{Latency (ms)} & \textbf{Size (MB)} \\
    \midrule
    \multirow{10}{7em}{ResNet-18} 
    & FP32 & 32/32 & 71.1 & 92.34 &  43.97\\
    & $LSQ_\psi$ & 8/8 & {\textbf{71.0}} 
 & 34.97 & 14.68 \\
    & $LSQ_\psi$ & 4/4 & 68.73 & 24.4 & 7.34\\
    & HAWQv2 & $4_{\text{MP}}/4_{\text{MP}}$  & 70.22 & 32.83 & 8.52\\
    & MCKP & $3_{\text{MP}}/4_{\text{MP}}$  & 69.66 & 28.05 & 7.66\\
    & DDQ & $4_{\text{MP}}/4_{\text{MP}}$ & \textbf{71.2} & 29.03  & 7.83\\
    & LIMPQ & $3_{\text{MP}}/3_{\text{MP}}$ & 69.7 & 20.03  & 7.55\\
    & LIMPQ & $4_{\text{MP}}/4_{\text{MP}}$ & \textbf{70.8} & 33.05  & 8.52\\
    & \textbf{\methodname{}} ($\beta_1$) & $2_{\text{MP}}/2_{\text{MP}}$& \textbf{70.87} & \textbf{9.3} & 7.87 \\
    & \textbf{\methodname{}} ($\beta_2$) & $2_{\text{MP}}/2_{\text{MP}}$  & 67.77 & \textbf{5.07} & 7.06\\
   \bottomrule
\end{tabular}
    \end{table}
     \begin{table}[H]\ContinuedFloat
\caption{{\em Cont.}} 
 
     \ \begin{tabular}{c l@{\hspace{0.705cm}} c@{\hspace{0.5cm}} c c c c}
    \toprule
    \textbf{Network} & \textbf{Method} & \textbf{Bitwidth} & \textbf{Acc (\%)} & \textbf{Latency (ms)} & \textbf{Size (MB)} \\
   \midrule
    \multirow{10}{7em}{ResNet-50}
    & FP32 & 32/32 & 80.1 & 263.64 & 102.06\\
    & $LSQ_{\psi}$ & 8/8 & \textbf{79.9} & 101.4 & 20.74 \\
    & $LSQ_{\psi}$ & 4/4 & 78.3 & 55.84 & 10.37 \\
    & $LSQ_\psi$ & 3/3 & 77.6 & \textbf{25.44} & 7.79\\
    & MCKP & $2_{\text{MP}}/4_{\text{MP}}$ & 75.28 & 46.42 & 7.96\\
    & HAQ & $3_{\text{MP}}/3_{\text{MP}}$ & 75.3 & --- &  ---\\
    & HAWQv2 & $2_{\text{MP}}/4_{\text{MP}}$ & 76.1 & 86.51 & 10.13 \\
    & LIMPQ & $3_{\text{MP}}/4_{\text{MP}}$ & 76.9 & \textbf{32.51}  & 8.11\\
    & \textbf{\methodname{}} ($\beta_1$) & $2_{\text{MP}}/2_{\text{MP}}$ & \textbf{79.23} & \textbf{37.52} & 11.89\\
    & \textbf{\methodname{}} ($\beta_2$) & $2_{\text{MP}}/2_{\text{MP}}$ & 78.5 & \textbf{34.47} & 7.75\\
   
       \midrule
     \multirow{10}{7em}{MobileNetV2}
     & FP32 & 32/32 & 71.80 & 104.34 & 17.86\\
     & $LSQ_{\psi}$ & 8/8 & 71.6 & 39.52 & 12.54\\
     & MCKP & $2_{\text{MP}}/8$ & 71.2 & 22.42 & 9.82\\
     & HAQ & $3_{\text{MP}}/3_{\text{MP}}$ & 66.99 & --- & --- \\
     & HAQ & $4_{\text{MP}}/4_{\text{MP}}$ & 71.47 & 15.89  & 10.47\\
     & DDQ & $4_{\text{MP}}/4_{\text{MP}}$ & \textbf{71.8} & 29.25  & 10.217\\
     & PROFIT & $4_{\text{MP}}/4_{\text{MP}}$ & 71.5 & ---  & ---\\
     & LSQ + BR & $3/3$ & 67.4 & \textbf{11.96}  & 11.429\\
     & \methodname{} ($\beta_1$) & $2_{\text{MP}}/2_{\text{MP}}$ & \textbf{71.29} & 15.01 & 6.34\\
     & \methodname{} ($\beta_2$) & $2_{\text{MP}}/2_{\text{MP}}$ & 71.2 & \textbf{11.85} & 10.12\\
    \bottomrule
    \end{tabular}

\end{table}



\begin{figure}[H]
	\includegraphics[width=.99\textwidth]{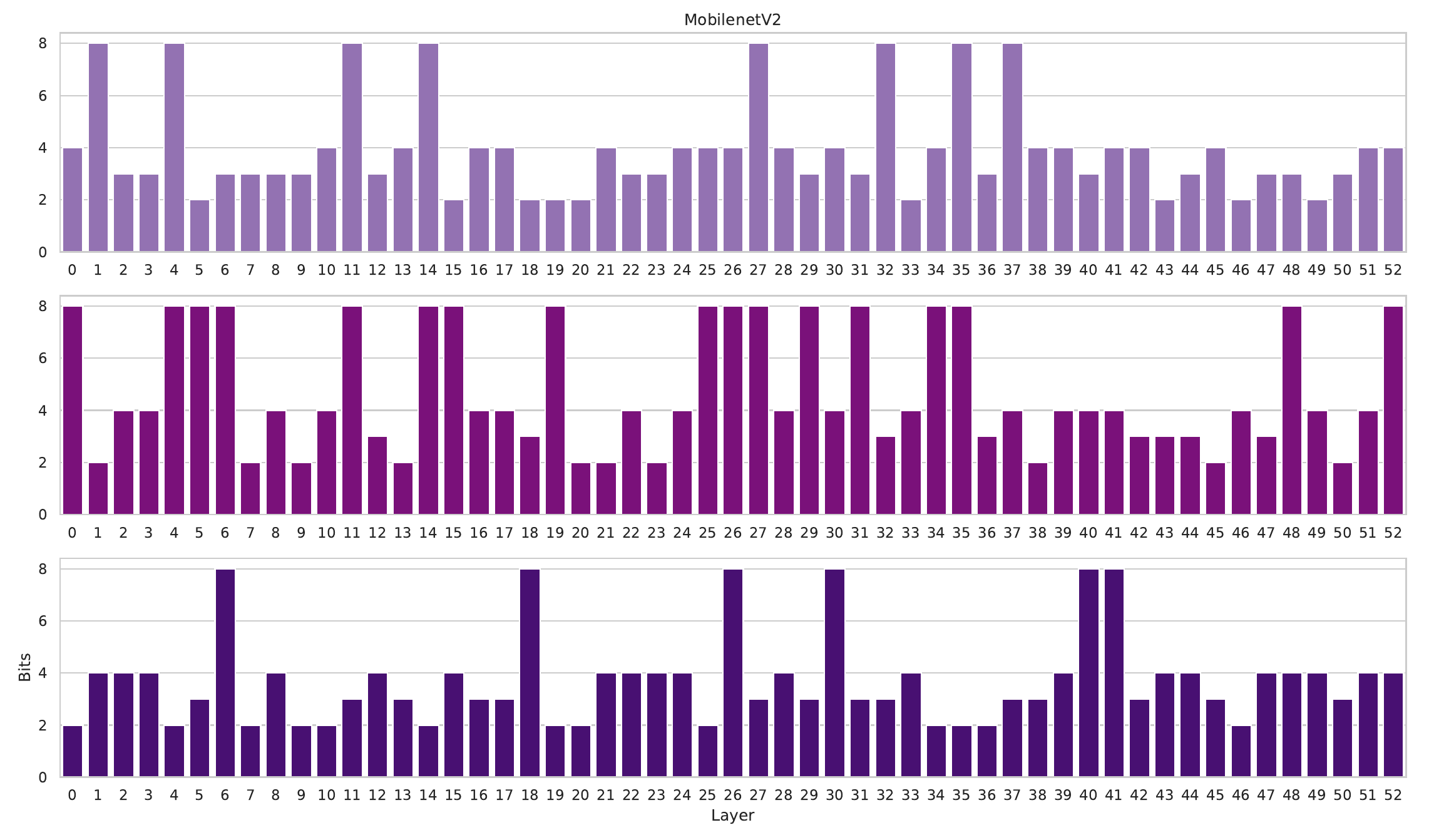} 
	\caption{{Quantization} 
 bit allocation of MobileNetV2 following our method using the simulator.
	The top figure is the SCALE-Sim setup; the~middle is the Eyeriss setup; the bottom is SCALE-Sim with low memory. Depthwise convolutions have a higher feature map and, thus, higher memory footprint, and~we can see that Algorithm \ref{alg:algo} allocates fewer bits when the system memory is low, i.e.,~the model is memory-bounded. Models with higher memory allocate the bits differently due to the locality of the boundary (memory or computational) by the layer.
	This figure does not include the first and last layers, which we quantize to 8 bits.}
	\label{fig:mobilebits}
\end{figure}

\begin{figure}[H]
	\includegraphics[width=.99\textwidth]{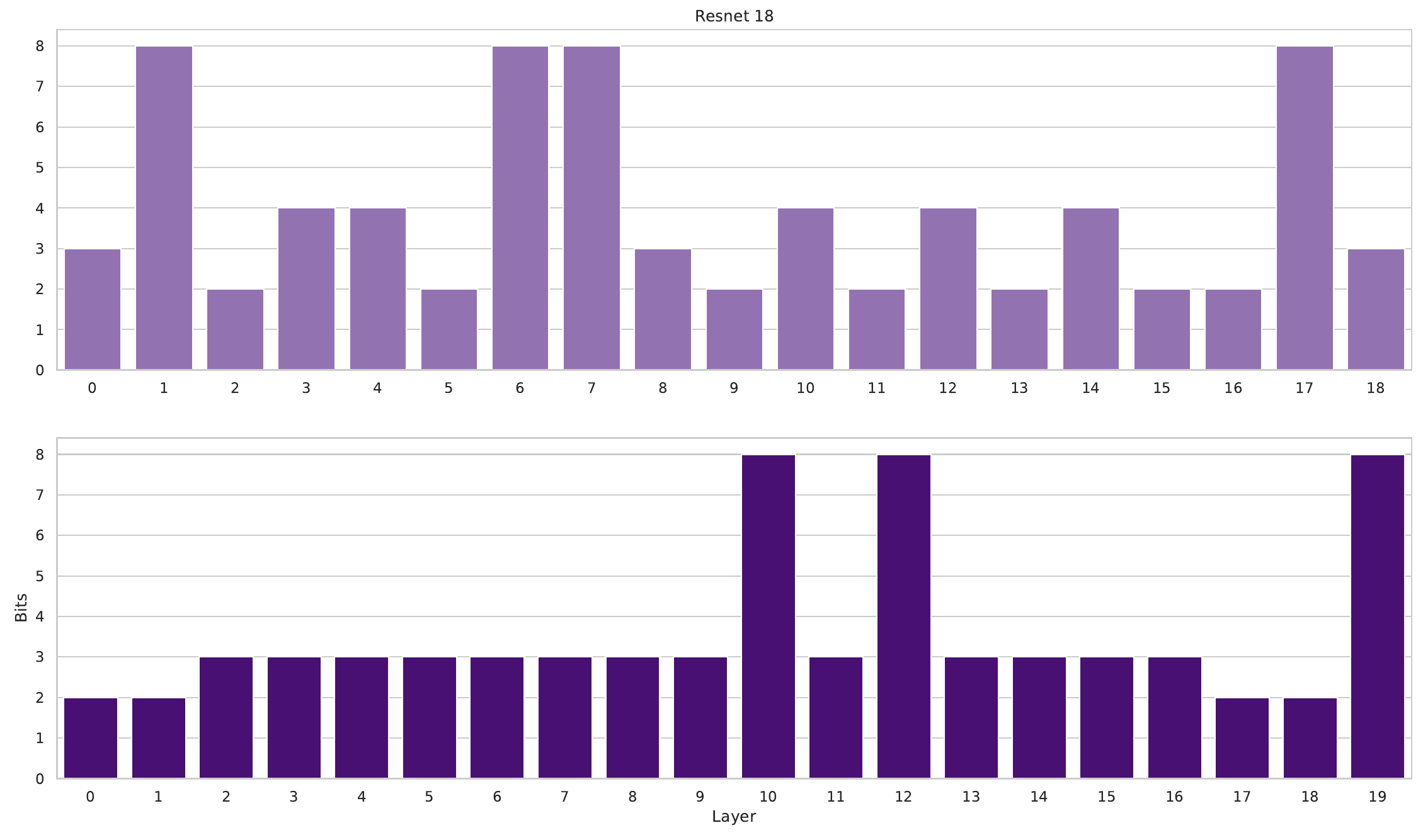} 
	\caption{{Quantization} 
 bit allocation of ResNet-18 following our method using the simulator.
	Both are for the SCALE-Sim setup. The~top figure is for $\beta=1$, and~the bottom is for $\beta=10$.
	When choosing a higher value for $\beta$, the~algorithm chooses lower precision for the model for the same hardware~constraint.}
	\label{fig:res18 bits}
\end{figure}

We tested our model on various hardware setups. {Figure} 
 \ref{fig:mobilebits} shows that our method produced a different bit allocation for MobilenetV2 for each hardware constraint. The~effect of different values for $\beta$ for the same model and the same hardware simulator is shown in Figure~\ref{fig:res18 bits}, and~other outcomes for different hardware constraints for ResNet-50 can be seen in Figure~\ref{fig:res50 bits}.

\vspace{-9pt}

\begin{figure}[H]
	\includegraphics[width=.9\textwidth]{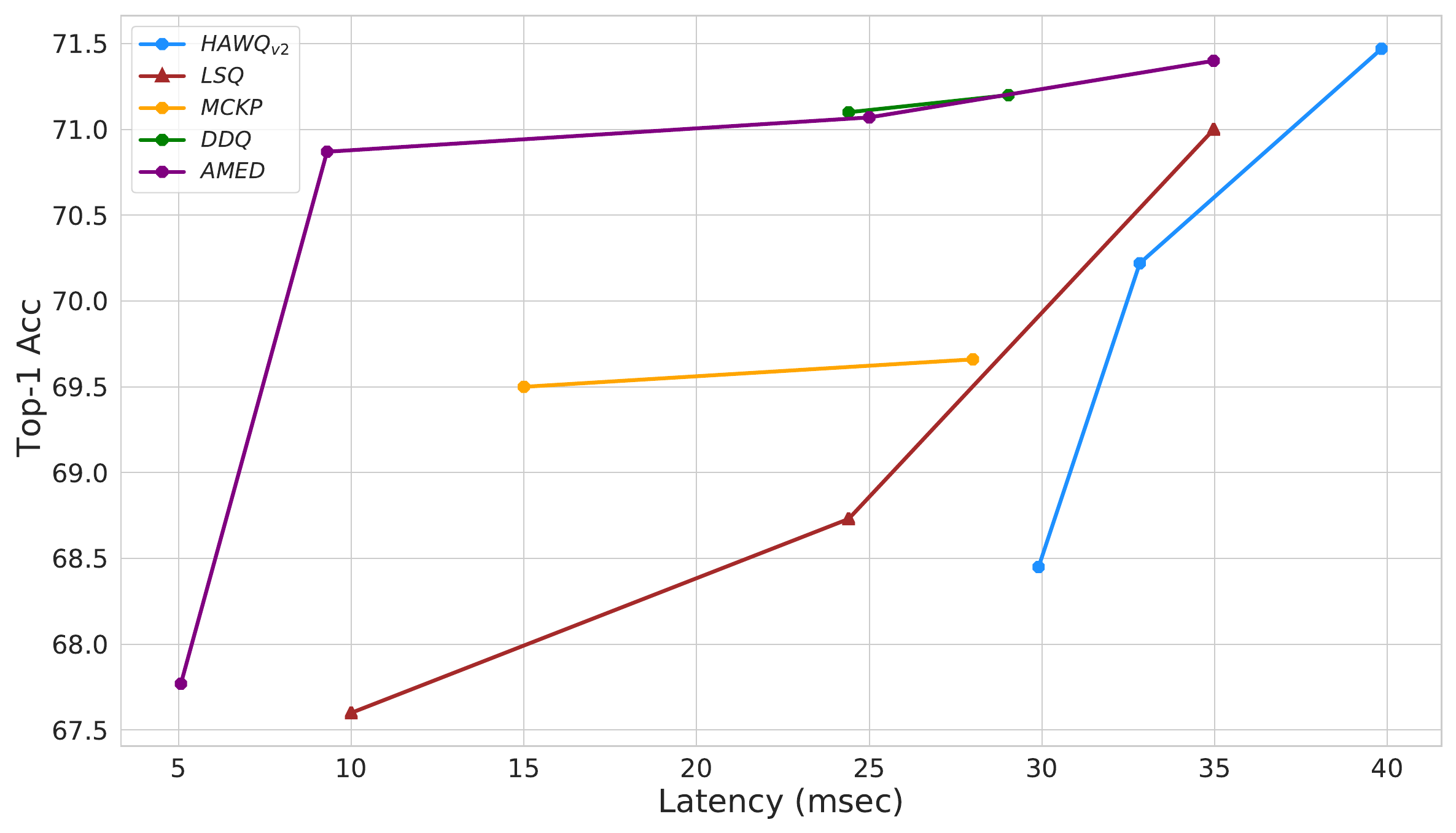} 
	\caption{ResNet-18 quantized models on a latency--accuracy plane. Circles are mixed precision, and triangles are uniform quantization.
	Our models achieved a better Pareto curve of the dominant solution in the two-dimensional plane for ultra-low precision.}
	\label{fig:resnet18}
\end{figure}

\begin{figure}[H]
	\includegraphics[width=.9\textwidth]{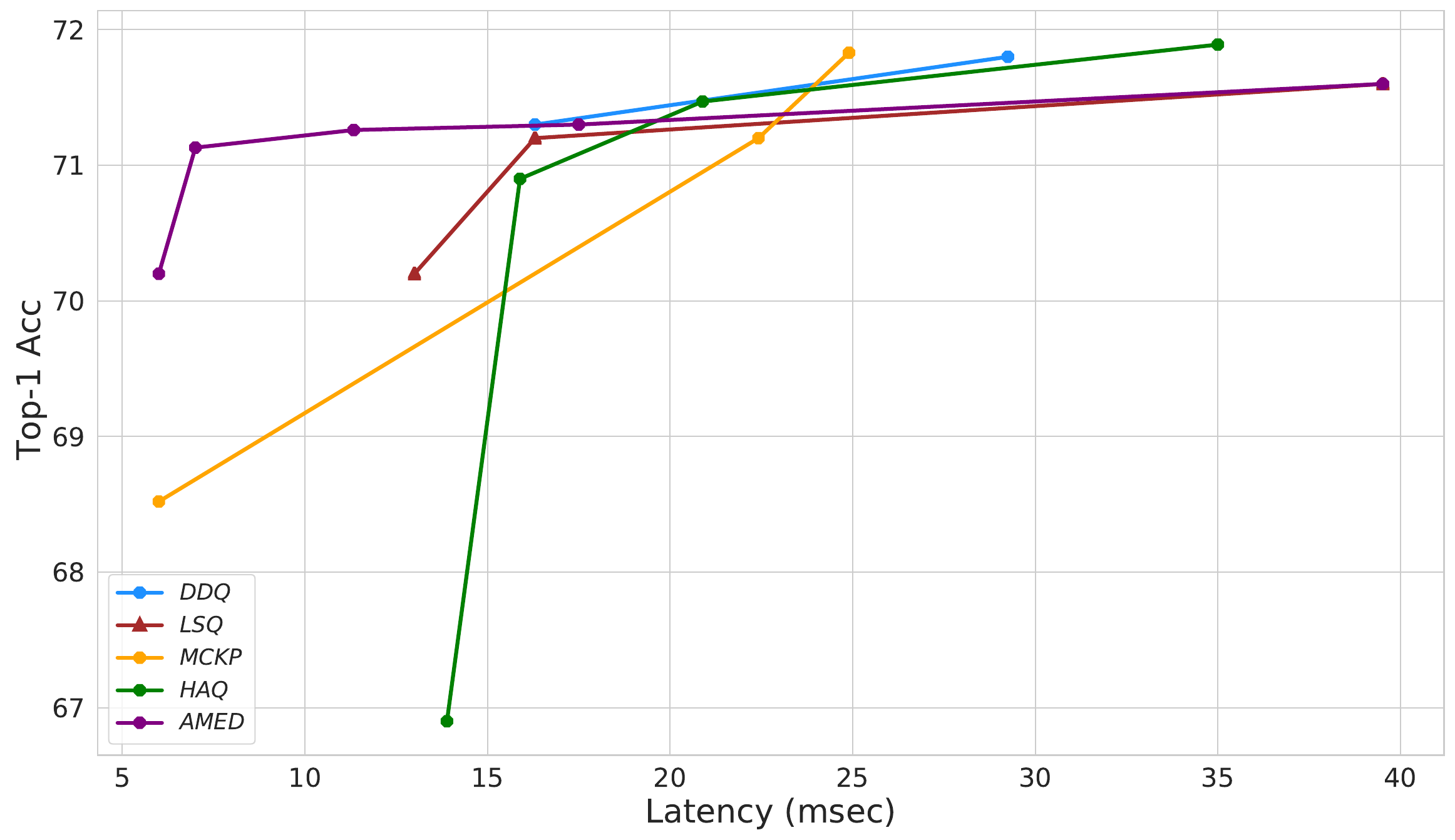} 
	\caption{MobileNetV2 quantized models on a latency--accuracy plane. Circles are mixed precision, and triangles are uniform quantization.
	Our models achieved a better Pareto curve of the dominant solution in the two-dimensional plane for ultra-low precision.}
	\label{fig:mobilev2}
\end{figure}

We report our quantized results in Table~\ref{tab:imagenet}. Because~\methodname{} can efficiently provide a trade-off between latency and accuracy, we can control the accuracy degradation and latency requirements easily by a simple adjustment of the hyperparameters. For~each architecture, we report multiple results that demonstrate this trade-off. As~seen in Table~\ref{tab:imagenet}, for~ResNet-18, we achieved more than a x2.6 latency improvement with only a 0.23\% drop in accuracy compared to the latest state-of-the-art quantization methods: LSQ~\cite{Esser2020LSQ} and DDQ~\cite{Zhang2021DifferentiableDQ}. For~ResNet-50, compared to the 4-bit models, we outperformed in accuracy  by 0.2--0.93\%  while still improving latency. For~MobileNetV2, we can also see more than a 1 ms improvement in latency compared to the HAQ~\cite{Wang2018HAQHA} 4-bit mixed-precision model with a 0.5\% degradation in accuracy. We emphasize that all results are prone to the $\beta$ setting and a degradation in accuracy can easily be compensated for by a higher latency. As~demonstrated in Figures~\ref{fig:resnet50}, \ref{fig:resnet18}, and \ref{fig:mobilev2}, \methodname{} achieves a better Pareto curve for the accuracy--latency trade-off, which can dominant other quantized~models.

\vspace{-6pt}
\unskip


\subsection*{Ablation~Study}
In this section, we describe our tests of \methodname{} on CIFAR100~\cite{cifar10} with a ResNet-18 architecture with different hyperparameters, as~listed in Table~\ref{tab:ablation}.
The method chooses the bitwidths, resulting in lower latency when $\beta$ is higher, as~expected, because~as shown in Equation~(\ref{eq:Z}), the~higher $\beta$ increases the objective's latency component. 
The use of EMA helps when we fix a small $\beta$ value. This result emphasizes  the importance of averaging the score of each bitwidth with older~samples.

\begin{table}[H]\setlength{\tabcolsep}{7.05mm}
   \caption{ Performance comparison of different hyperparameters of ResNet-18 on CIFAR100. We used the SCALE-Sim simulator described in Table~\ref{tab:HW setup}, and~the latency is normalized to one image~inference.}
    \label{tab:ablation}
    \begin{tabular}{c c c c c}
    \toprule
    \boldmath{$\beta$} & \textbf{EMA}  & \textbf{Top-1 (\%)} & \textbf{Top-5 (\%}) &\textbf{Latency (ms})\\
    \midrule
    1 & 0.9 & 75.59 & 96.15 & 10.14\\
     1 & 0.5 & 73.08 & 92.19 & 14.68\\
     1 & 0.2 & 77.86 & 95.09 & 14.68\\
     1 & 0.1 & 77.58 & 92.97 & 12.47\\
     1 & 0.01 & 78.21 & 94.97 & 8.99\\
    100 & 0.01 & 77.29 & 91.44 & 6.15\\
     10 & 0.01 & 78.29 & 93.97 & 8.15\\
     1  & 0.01 & 78.52 & 94.53 & 8.98\\
     0.1 & 0.01 & 78.72 & 96.23 & 12.61\\
     
    \bottomrule
    \end{tabular}
 
\end{table}


\section{Discussion}

\label{sec:conclusions}
In this paper, we introduced a novel mixed-precision quantization method called \methodname{}. 
The proposed method relies on a novel meta-bit allocation strategy that finds an optimal bitwidth among different neural network layers by measuring direct signals from a hardware accelerator simulator. 
Our method significantly reduces the required exploration space compared to previous mixed-precision methods, due to the simplicity of the perspective of the low-degree objective.
The extensive evaluations we performed demonstrated the superiority of our method over standard image classification benchmarks in terms of the accuracy--latency trade-off compared to the prior state of the art. The~ability to obtain higher accuracy in a shorter training time results in lower time-to-market~solutions.

Future work to examine this method for efficient NAS could yield a computationally efficient search, which will reduce the carbon footprint of the procedure and reveal better models in terms of inference~time.

Another intriguing future topic is the exploration of different loss terms, solving dense prediction tasks like semantic segmentation~\cite{chen2018encoderdecoder,kimhi2023semisupervised}.

Improvements in the simulator, such as enabling dynamic workflow based on the computational graph, or~support for special hardware solutions such as fast sparse matrix  multiplication~\cite{Srivastava2020MatRaptorAS}, combined with pruning, could result in very promising~outcomes.

\subsection*{{Future} 
 Directions}

{{Computationally efficient NAS:}} 
~Integrating \methodname{} with efficient Neural Architecture Search (NAS) techniques has the potential to yield a significantly more computationally efficient search process. This would not only reduce the carbon footprint associated with the search procedure, but also potentially uncover models with superior inference~times.

{{Exploration of diverse loss terms:}} The ability to directly measure various empirical values within the deployment system through the hardware simulator opens doors for exploring a much broader range of loss terms. Unlike previous approaches, differentiability is no longer a prerequisite for loss terms, allowing us to consider factors like power consumption, memory usage, and~bandwidth within the existing deployment~environment.

{{Enhanced Simulator Capabilities:}} Further improvements to the simulator, such as enabling dynamic workflows based on the computational graph or incorporating support for specialized hardware solutions like Fast Sparse Matrix Multiplication~\cite{Srivastava2020MatRaptorAS} alongside pruning techniques, hold immense promise for achieving groundbreaking~results.






\vspace{6pt} 




\authorcontributions{ 
Conceptualization and methodology, M.K, C.B., and A.M.; code, M.K.;  formal analysis, writing and visualization, M.K., T.R., C.B., and A.M.; resources, C.B. and A.M.; supervision and Funding, A.M. All authors have read and agreed to the published version of the~manuscript.}

\funding{This research was funded by Israel Innovation Authority, Nofar~grant.}

\dataavailability{Accessed data for the experiments for CIFAR-10 and ImageNet can be found at \url{https://www.cs.toronto.edu/~kriz/cifar.html} and \url{https://www.image-net.org/} (accessed on 28 January 2023), respectively.} 




\acknowledgments{The authors would like to thank Tal Kopetz and Olya Sirkin from CEVA LTD for the helpful discussions and brainstorming during this~project.}

\conflictsofinterest{The authors declare no conflicts of~interest.}



\abbreviations{Abbreviations}{
The following abbreviations are used in this manuscript:\\

\noindent 
\begin{tabular}{@{}ll}
FLOP & floating-point operations\\
MAC & multiply--accumulate \\
DNNs & deep neural networks\\
MP & mixed-precision\\
FP32 & floating-point with 32 bits\\
NAS & Neural Architectural Search\\
CE & cross-entropy\\
PTQ & Post-Training Quantization\\
QAT & Quantization-Aware Training \\
PE & processing element \\
SRAM & static random-access memory\\

\end{tabular}
}

\appendixtitles{yes} 

\appendixstart
\appendix




\section{Additional~Experiments}
This Appendix includes the results for the Eyeriss hardware setup, as~listed in Table~\ref{tab:HW setup}.
The results are presented in Table~\ref{tab:imagenet hw3} in this Appendix and show that our method finds a different precision per layer for each model.
We also found that, for a low memory boundary, our model chooses strategic layers to maintain high precision regardless of the size of the feature map, and~tries to compensate for the latency over other layers.
More results and more experiments can be reproduced using the GitHub repository mentioned in this~paper.

\begin{table}[H]\setlength{\tabcolsep}{2.65mm}
    \caption{{The} 
 same comparison as in Table~\ref{tab:imagenet} and the same notations, but~with the latency from the simulator of the Eyeriss setup from Table~\ref{tab:HW setup}. Note that our model yields a different bitwidth for each layer because the signal from the hardware is different in this~setup. }
    \label{tab:imagenet hw3}
    \begin{tabular}{c c c c c c}
    \toprule
    \textbf{Network} & \textbf{Method} & \textbf{Bitwidth} & \textbf{Acc (\%)} & \textbf{Latency (ms)} & \textbf{Size (MB)} \\
    \midrule
    \multirow{8}{7em}{ResNet-18} 
    & FP32 & 32/32 & 71.1 & ---  & 43.97\\
    & $LSQ_\psi$ & 8/8 & 70.0 & 3.42 & 14.68 \\
    & $LSQ_\psi$ & 4/4 & 68.73 & 0.85 & 7.34\\
    & HAWQv2 & $4_{MP}/4_{MP}$  & 70.22 & 2.94 & 8.52\\
    & MCKP & $3_{MP}/4_{MP}$  & 69.66 & 2.13 & 7.66\\
    & DDQ & $4_{MP}/4_{MP}$ & 71.2  & 2.22  & 7.83\\
    & \methodname{} & $2_{MP}/2_{MP}$  & 70.84  & {{0.32}} 
 & 6.16 \\
    & \methodname{} & $2_{MP}/2_{MP}$  & 71.0 & {0.55} & 6.6\\
    \midrule
    \multirow{8}{7em}{ResNet-50}
    & FP32 & 32/32 & 80.1 & --- & 102.06\\
    & $LSQ_{\psi}$ & 8/8 & 79.9 & 9.65 & 20.74 \\
    & $LSQ_{\psi}$ & 4/4 & 78.3 & 2.42 & 10.37 \\
    & $LSQ_\psi$ & 3/3 & 77.6 & 0.8 & 7.79\\
    & MCKP & $2_{MP}/4_{MP}$ & 75.28 & 3.22 & 7.96\\
    & HAWQv2 & $2_{MP}/4_{MP}$ & 76.1 & 7.82 & 10.13 \\
    & {\methodname{}} & $2_{MP}/2_{MP}$ & {79.34} & {3.26} & 9.74\\
    & {\methodname{}} & $2_{MP}/2_{MP}$ & {79.43} & {3.45} & 9.75\\
    \midrule
     \multirow{10}{7em}{MobileNetV2}
     & FP32 & 32/32 & 71.80 & --- & 17.86\\
     & $LSQ_{\psi}$ & 8/8 & 71.6 & 7.8 & 12.54\\
     & MCKP & $2_{MP}/8$ & 71.2 & 4.44 & 9.82\\
     & HAQ & $3_{MP}/3_{MP}$ & 70.9 & --- & --- \\
     & HAQ & $4_{MP}/4_{MP}$ & {71.47} & {2.03}  & 10.47\\
     & DDQ & $4_{MP}/4_{MP}$ & 71.8 & 5.87  & 10.217\\
     & PROFIT & $4_{MP}/4_{MP}$ & 71.5 & ---  & ---\\
     & LSQ + BR & $3/3$ & 67.4 & 3.9  & 11.429\\
     & \methodname{} & $2_{MP}/2_{MP}$ & {71.24} & {1.56} & 12.1\\
     & \methodname{} & $2_{MP}/2_{MP}$ & {70.97} & {1.21} & 9.53\\
    \bottomrule
    \end{tabular}

\end{table}
\unskip

\section{Bit~Allocations}
\label{app:bit allocations}
\begin{figure}[H]
	\includegraphics[width=1.\textwidth]{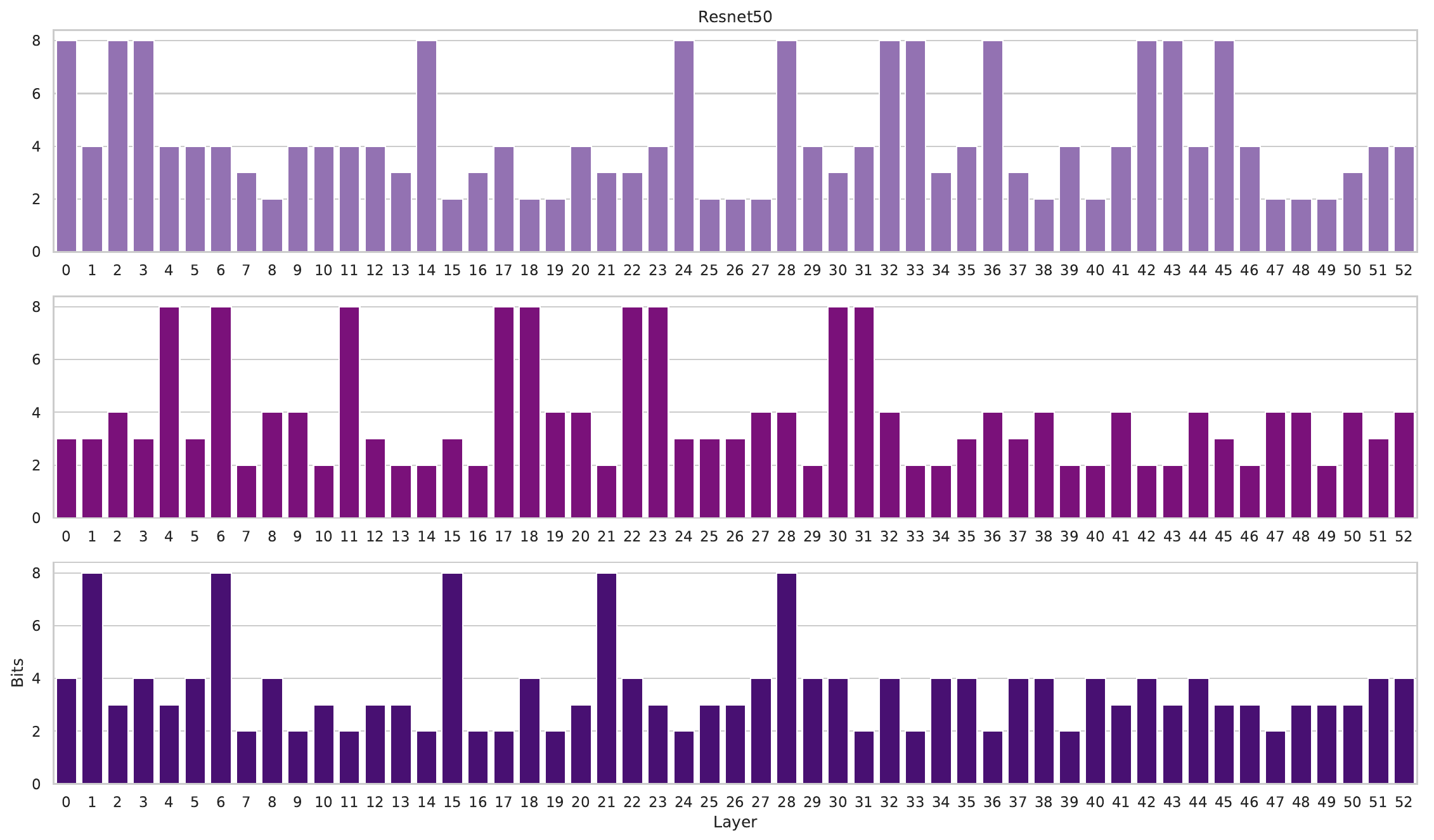} 
	\vspace{-0.3cm}
	\caption{Quantization bit allocation of ResNet-50 following our method using the simulator.
	The top figure is the SCALE-Sim setup; the~middle is the Eyeriss setup;~the bottom is SCALE-Sim with low memory.}
	\label{fig:res50 bits}
\end{figure}

In Figure~\ref{fig:res50 bits}, we add the bit allocation for ResNet-50 with a different hardware setup, and~in Figure~\ref{fig:res18 bits}, we use different values for $\beta$ for~ResNet-18.

One can see that both affect the bit allocation significantly towards building a more efficient network when the memory boundary is closer or when we increase the hardware constraints on the~objective.

\vspace{-6pt}
\begin{figure}[H]

\begin{adjustwidth}{-\extralength}{0cm}
\centering 
\includegraphics[width=1.\textwidth]{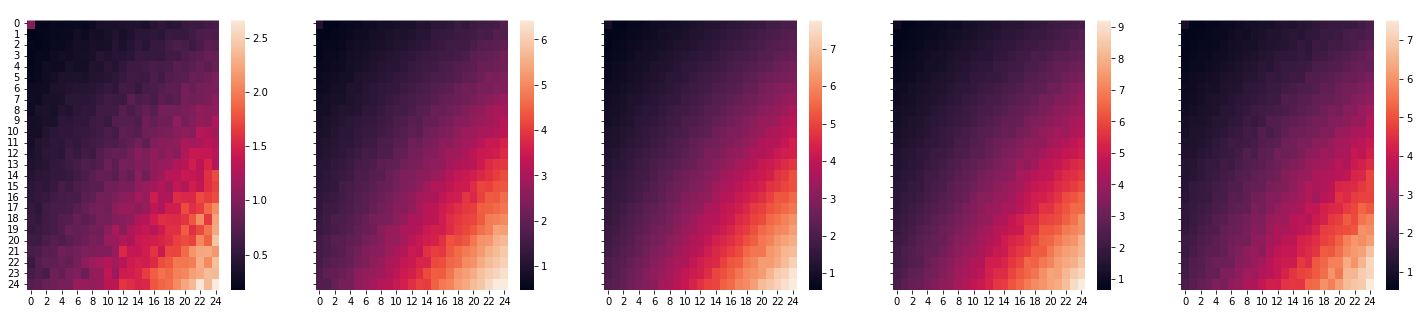} 
\end{adjustwidth}
	\caption{{Visualization} 
 of the loss surface of two subsequent layers of
ResNet-18. At~a higher bitwidth (left), the~interactions between layers are relatively small, making layerwise optimization possible. On~the other hand, with a~bitwidth decrease (right), with~an increase in the quantization loss, the~interactions become tangible and the loss is higher. A~per-layer optimization  depends on the initial point and is potentially sub-optimal.}
	\label{fig:LOSSES}
\end{figure}
\unskip

\section{Reports}
\label{app:reports}
The simulator provides the following detailed reports about the performance during the inference of a~CNN:
\begin{itemize}
    \item {{Computation report} 
}: Provides layerwise details about Total Cycles, Stall Cycles, Overall Utilization, Mapping Efficiency, and~Computation Utilization. An~example is shown in Table~\ref{tab:Compute report}.
    \item{Bandwidth report}: Provides layerwise details about Average IFMAP SRAM Bandwidth, Average FILTER SRAM Bandwidth, Average OFMAP SRAM Bandwidth, Average IFMAP DRAM Bandwidth, Average FILTER DRAM Bandwidth, and~Average OFMAP DRAM Bandwidth.
    \item {Detailed access report}: Provides layerwise details about the number of reads and writes, and~the start and stop cycles, for~both of the above-mentioned reports.
\end{itemize}

\begin{table}[H]\setlength{\tabcolsep}{6.05mm}
 \caption{An example of the SCALE-Sim simulator computation report similar to~\cite{MAX78000} for MobileNetV2 uniformly quantized to 2 bits (not including the FC layer).
    }
    \label{tab:Compute report}
\small
\begin{adjustwidth}{-\extralength}{0cm}
\centering 
\begin{tabular}{llllll}
\toprule   
\textbf{LayerID} & \textbf{Total}   \textbf{Cycles} & \textbf{Stall}  & \textbf{Overall Util} \% & \textbf{Mapping Efficiency \%} & \textbf{Compute Util \%} \\
\midrule   
0& {137,148} 
& 0& 16.10753928     & 53.09483493& 16.10742183     \\
1& {132,649}& 0& 43.61369102     & 53.00234993& 43.61336223     \\
2& {36,847}& 0& 9.574727929     & 28.125& 9.574468085     \\
3& 61,151& 0& 15.70538503     & 76.5625& 15.70512821     \\
4& 701,907& 0& 71.27146118     & 76.38573961& 71.27135964     \\
5& 15,483& 0& 24.68513854     & 40.625& 24.6835443      \\
6& 25,283& 0& 21.22176957     & 76.04166667& 21.22093023     \\
7& 374,807& 0& 71.87994421     & 75.31844429& 71.87975243     \\
8& 20,187& 0& 28.399465       & 40.625& 28.39805825     \\
9& 25,283& 0& 21.22176957     & 76.04166667& 21.22093023     \\
10& 374,807& 0& 71.87994421     & 75.31844429& 71.87975243     \\
11& 5149& 0& 36.4002719      & 52.0625& 36.39320388     \\
12& 9399& 0& 25.28460475     & 74.265625& 25.28191489     \\
13& 157,519& 0& 70.24724002     & 72.76722301& 70.24679406     \\
14& 6349& 0& 39.36052922     & 52.0625& 39.35433071     \\
15& 9399& 0& 25.28460475     & 74.265625& 25.28191489     \\
16& 157,519& 0& 70.24724002     & 72.76722301& 70.24679406     \\
17& 6349& 0& 39.36052922     & 52.0625& 39.35433071     \\
18& 9399           & 0& 25.28460475     & 74.265625             & 25.28191489     \\
19& 157,519& 0& 70.24724002     & 72.76722301& 70.24679406     \\
20& 3555& 0& 34.11392405     & 45.1171875& 34.10433071     \\
21& 6173& 0& 38.2998542      & 75.390625& 38.29365079     \\
22& 123,129& 0& 76.17864191     & 77.54464286& 76.17802323     \\
23& 6243& 0& 38.8515137      & 45.1171875& 38.84529148     \\
24& 6173& 0& 38.2998542      & 75.390625& 38.29365079     \\
25& 123,129& 0& 76.17864191     & 77.54464286& 76.17802323     \\
26& 6243& 0& 38.8515137      & 45.1171875& 38.84529148     \\
27& 6173& 0& 38.2998542      & 75.390625& 38.29365079     \\
28& 123,129& 0& 76.17864191     & 77.54464286& 76.17802323     \\
29& 6243& 0& 38.8515137      & 45.1171875& 38.84529148     \\
30& 6173& 0& 38.2998542      & 75.390625& 38.29365079     \\
31& 123,129& 0& 76.17864191     & 77.54464286& 76.17802323     \\
32& 6243& 0& 57.68861124     & 66.9921875& 57.6793722      \\
33& 11,059& 0& 48.01858215     & 79.0234375& 48.01424051     \\
34& 262,299& 0& 80.32093146     & 81.28125& 80.32062524     \\
35& 8931& 0& 60.48874706     & 66.9921875& 60.48197492     \\
36& 11,059& 0& 48.01858215     & 79.0234375& 48.01424051     \\
37& 262,299& 0& 80.32093146     & 81.28125& 80.32062524     \\
38& 8931& 0& 60.48874706     & 66.9921875& 60.48197492     \\
39& 11,059& 0& 48.01858215     & 79.0234375& 48.01424051     \\
40& 262,299& 0& 80.32093146     & 81.28125& 80.32062524     \\
41& 3827& 0& 58.33714398     & 64.59960938& 58.32190439     \\
42& 7103& 0& 51.84649092     & 71.92687988& 51.83919271     \\
43& 139,231& 0& 72.87237576     & 73.39477539& 72.87185238     \\
44& 6131& 0& 60.69054803     & 64.59960938& 60.68065068     \\
45& 7103& 0& 51.84649092     & 71.92687988& 51.83919271     \\
46& 139,231& 0& 72.87237576     & 73.39477539& 72.87185238     \\
47& 6131& 0& 60.69054803     & 64.59960938& 60.68065068     \\
48& 7103& 0& 51.84649092     & 71.92687988& 51.83919271     \\
49& 139,231& 0& 72.87237576     & 73.39477539& 72.87185238     \\
50& 12,263& 0& 60.31099649     & 64.20084635& 60.30607877     \\
51& 16,043& 0& 61.18127844     & 73.03059896& 61.1774651      \\
52& 21,471& 0& 2.916724885     & 3.057861328& 2.916589046    \\
\bottomrule
\end{tabular}
\end{adjustwidth}
   
\end{table}


\newpage
\begin{adjustwidth}{-\extralength}{0cm}

\reftitle{References}

\PublishersNote{}
\end{adjustwidth}
\end{document}